\documentclass[sigconf]{acmart}

\begin{document}

\title{DifCluE: Generating Counterfactual Explanations with Diffusion Autoencoders and modal clustering}


\author{Suparshva Jain}
\affiliation{%
  \institution{TCS Research}
  \city{Delhi}
  \country{India}}
\email{Suparshva.Jain@tcs.com}

\author{Amit Sangroya}
\affiliation{%
  \institution{TCS Research}
  \city{Delhi}
  \country{India}}
  
\email{Amit.Sangroya@tcs.com}

\author{Lovekesh Vig}
\affiliation{%
  \institution{TCS Research}
  \city{Delhi}
  \country{India}}
\email{Lovekesh.Vig@tcs.com}

% \renewcommand{\shortauthors}{Sangroya et al.}

\begin{abstract}
  Generating multiple counterfactual explanations for different modes within a class presents a significant challenge, as these modes are distinct yet converge under the same classification. Diffusion probabilistic models (DPMs) have demonstrated a strong ability to capture the underlying modes of data distributions. In this paper, we harness the power of a Diffusion Autoencoder to generate multiple distinct counterfactual explanations. By clustering in the latent space, we uncover the directions corresponding to the different modes within a class, enabling the generation of diverse and meaningful counterfactuals. We introduce a novel methodology, \textit{DifCluE}, which consistently identifies these modes and produces more reliable counterfactual explanations. Our experimental results demonstrate that \textit{DifCluE} outperforms the current state-of-the-art in generating multiple counterfactual explanations, offering a significant advancement in model interpretability.
  
\end{abstract}

%    <concept>
%        <concept_id>10010147.10010257.10010293.10003660</concept_id>
%        <concept_desc>Computing methodologies~Classification and regression trees</concept_desc>
%        <concept_significance>100</concept_significance>
%        </concept>
%  </ccs2012>

% \ccsdesc[100]{Computing methodologies~Classification and regression trees}


\maketitle

\section{Introduction}

\begin{figure*}[h]
    \centering
    \includegraphics[trim={10cm 0 10cm 0},scale=0.6]{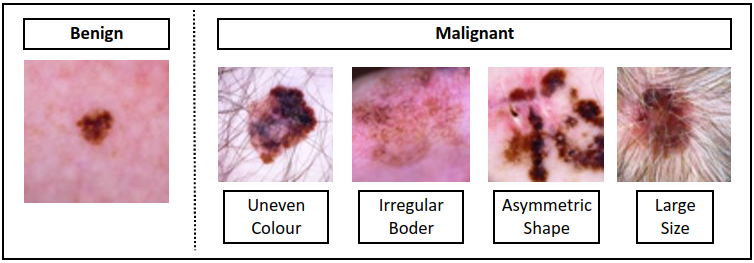}
    \caption{Various features that melanoma skin lesions may exhibit as compared to a non melanoma skin lesions}
    \label{fig:fig_melanoma}
\end{figure*}

Real-world classification problems often  deal with situations in which the feature set can have multiple modes but are classified within a single class. For example, a skin lesion can be classified as malignant due to many different value combinations of clinical features like a asymmetry, size, uneven colour, irregular borders etc (Figure~\ref{fig:fig_melanoma}). In order to achieve larger trust in the outcomes of deep models, we often rely on counterfactual explanations. Generating counterfactual explanations in such a case can prove to be tricky, since the class can have multiple modes. This challenge that has received limited attention. Prior work on counterfactual generation utilizes three distinct approaches; (a) generate multiple counterfactuals based on a distinctness loss \cite{DBLP:journals/corr/abs-2105-15164}, (b) use semantic information gathered from an encoder to generate counterfactuals \cite{downs2020cruds}, and (c) use gradient information from the target class for counterfactual explanations of input instances \cite{jeanneret2022diffusion}.

However, during the generation of counterfactuals, when there are multiple modes in a class, the generation method needs to generate counterfactual exemplars for  each mode with in the class. This is important to get the coverage of all valid feature perturbations that would classify an input into a target class, and helps us to get a better understanding of the target class. Additionally, being able to generate perturbations at the conceptual feature-level, and not just at the pixel-level or at the image-attribution level, would improve counterfactual exemplars and in-turn improve the level of explainability of the counterfactual examples for the end-users. 

\begin{figure*}[hbt]
    \centering
    \includegraphics[scale=0.6]{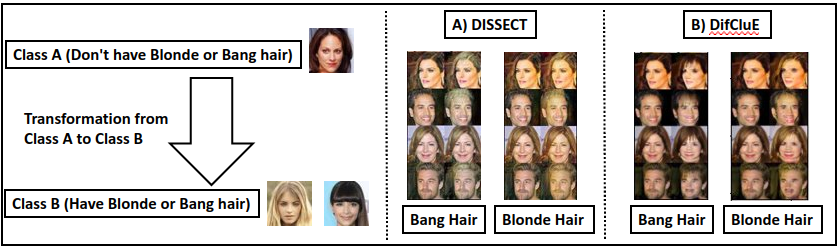}
    \caption{Comparison between the counterfactual explanations provided by DISSECT model and DifCluE model }
    \label{fig:fig_diffae_res}
\end{figure*}

In our work, we explore the ability of a Diffusion Autoencoder to come up with multiple distinct counterfactual explanations by performing clustering in latent space, when we have a class which contains more than one mode. We show that clustering in the latent space can help us to uncover the various modes within a class helping us to generate multiple distinct counterfactual explanations.  Diffusion probabilistic models (DPMs) have been shown to outperform GANs in terms of the quality of the images generated. In this paper, we make following key contributions:

\begin{enumerate}
    \item We propose our methodology \textit{DifCluE} (\textbf{Dif}fuse, \textbf{Clu}ster and \textbf{E}xplain), that generates distinct counterfactual explanations. This approach outperforms the current state of the art models for various quality parameters such as modal coverage, and quality of images generated.
    
    % \item We further used gradient based perturbations to generate images corresponding to a particular mode, as opposed to single step image generation used by prior approaches. 
    
    \item Our results demonstrate superior performance in terms of mode capture and counterfactual generation when compared to prior approaches (See Figure~\ref{fig:fig_diffae_res}).

    \item We demonstrate that \textit{DifCluE} approach which is based on a diffusion autoencoder to capture the data distribution, and clustering the resulting class embeddings leads to a better discovery of intra-class modes. Our experimental results show that after mixing different classes into a single class and employing clustering we are able to closely recover the original classes, suggesting that some degree of  disentanglement is being achieved in the intra class modes.

\end{enumerate}

\section{Related Work}

Counterfactual explanation generation using diffusion models represents a cutting-edge approach in interpretable AI. Diffusion models, known for their ability to generate high-quality samples, are now being adapted to create counterfactuals—hypothetical scenarios that explain model decisions by altering input features. By leveraging the stochastic nature of diffusion processes, these models can produce diverse and plausible counterfactuals, enhancing the transparency and robustness of AI systems in fields such as finance, healthcare, and legal decision-making.

Diffusion probabilistic models (DPM) have achieved superlative image generation quality \cite{DBLP:journals/corr/Sohl-DicksteinW15,DBLP:journals/corr/abs-1907-05600,DBLP:journals/corr/abs-2006-11239,DBLP:journals/corr/abs-2011-13456} and  Diffusion-based image editing has drawn much attention. There are two main categories of image editing and generation. Firstly, image-guided generation works to edit an image by mixing the latent variables of DPM and the input image \cite{DBLP:journals/corr/abs-2108-02938,DBLP:journals/corr/abs-2201-09865,DBLP:journals/corr/abs-2108-01073}. However, using images to specify the attributes for editing may cause ambiguity, as pointed out by \cite{kwon2023diffusion} . Secondly, the classifier guided works \cite{DBLP:journals/corr/abs-2105-05233, DBLP:journals/corr/abs-2111-14818, DBLP:journals/corr/abs-2112-05744} edit images by utilizing the gradient of an extra classifier. In our work, we build upon the second category, aiming to generate multiple counterfactual explanations with the the help of clustering and gradients of the classifier.

Bengio et al. introduced disentangled representation learning, where the target is to discover underlying explanatory factors of the observed data~\cite{DBLP:journals/corr/abs-1206-5538}. The disentangled representation is defined such that each dimension (or set of dimensions) of the representation correspond to an independent factor. Based on this definition, some VAE-based techniques achieve disentanglement  by constraining the probabilistic distributions of representations\cite{chen2018isolating,kim2018disentangling,higgins2016beta}. In \cite{DBLP:journals/corr/abs-1811-12359}, authors point out the identifiable problem by proving that only these constraints are insufficient for disentanglement and that extra inductive bias is required. 

A number of work has been done in generation of counterfactual explanations. Nemirovsky et al. uses a Residual Generative Adversarial Network (RGAN) to generate counterfactuals~\cite{DBLP:journals/corr/abs-2009-05199}. They enhance a regular GAN output with a residual that appear like perturbations used in counterfactual search, and then use a fixed target classifier to provide the counterfactuals. Support Vector Data description based counterfactual generation is described in~\cite{Carlevaro2023}. They use data envelopes extracted via singular value decomposition to generate counterfactuals for multi-class setting. However, different modes in a class does not seem to be addressed. In \cite{sun2023inherently}, authors generate class-specific attribution maps based on counterfactuals to find which region of the image are important for that classification and then use a simple logistic regression classifier to make predictions. Authors claim these methods to be inherently interpretable. In \cite{barberan2022neuroviewrnn} authors describe NeuroView-RNN as a family of new RNN architectures that explains how each hidden state per time step contributes to the decision-making process in a quantitative manner. Each member of the family is derived from a standard RNN architecture by concatenation of the hidden steps into a global linear classifier, so the weights of the classifier have a linear mapping to the hidden states and thus interpretability improves.

In \cite{jeanneret2022diffusion}, authors use gradient information from the target class for counterfactual generation of input instances to guide the generation process, instead of using one trained on noisy instances. They show that this approach is better than other methods. \cite{kanehira2019multimodal} have  proposed a method to generate counterfactual explanations with multimodal information such as visual and textual description information. They generate counterfactual explanations by training a classification model, which is the target of the explanation, and by training an auxiliary explanation model in a post-hoc manner by utilizing output and mid-level activation of the target classifier after freezing its weights to prevent change in output. 

Unfortunately, none of the above methods seem to address counterfactual explanation generation for multiple modes in a class. Our proposed \textit{DifClue} approach utilizes K-means clustering on the latent space obtained using Diffusion Autoencoder \cite{preechakul2022diffusion} to leverage the resulting disentanglement and generate counterfactuals via perturbations in the directions of the modes. We aim to use the realism as a measure to evaluate whether the generated images could have come from the original image set or not~\cite{DBLP:journals/corr/HeuselRUNKH17}.

% \begin{figure*}[t]
%     \centering
%     \includegraphics[scale=0.65]{Submission-2024/LaTeX/dissect_archi.png}
%     \caption{Overall Structure of DISSECT Model.}
%     \label{fig:fig_flowchart_dissect}
% \end{figure*}

\begin{figure*}[t]
    \centering
    \includegraphics[scale=0.6]{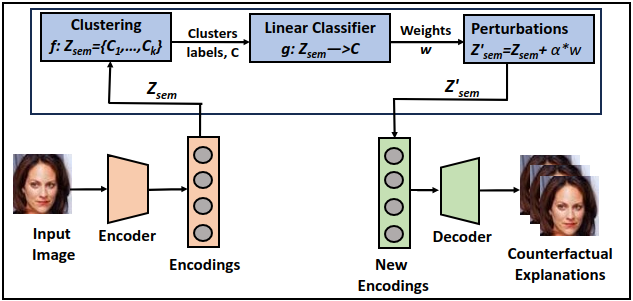}
    \caption{Overall flowchart for generating multiple counterfactual explanation using DifCluE}
    \label{fig:fig_flowchart}
\end{figure*}
% \vspace{-2em}

\section{Methodology}

% The idea behind DifCluE is that it helps us to identify and generate counterfactual explanations for classes which have more than one mode. This results in better understanding of the class in question and provides explainabilty. We would therefore like to compare it with current state of the art model and also evaluate its performance. 

The core concept of \textit{DifCluE} is its ability to identify and generate counterfactual explanations for classes with multiple modes, offering a deeper understanding of complex class structures. This approach enhances explainability by providing insights into the varied conditions that define a class. To assess its effectiveness, we aim to compare DifCluE with existing state-of-the-art models and thoroughly evaluate its performance.

\subsection{DifCluE Approach}

%  Diffusion Autoencoder has been shown to capture most of the important features of an image in the encodings, while using a Diffusion Probabilistic Model (DPM) to modelling the remaining stochastic variations. The encoding were shown to be very useful for classification even by a linear classifier. We exploit these encodings to generate multiple counterfactual explanations. Generating multiple counterfactual explanations helps us to better explain the internal variations within a class.
Our \textit{DifCluE} approach is designed to generate multiple distinct counterfactual explanations, starting with a Diffusion Autoencoder that serves as the foundation. We first train an Encoder-Decoder model based on a diffusion process, which effectively captures the key features of input data in its latent space. This latent space encodes the essential characteristics of each data point, providing a rich and compact representation that is crucial for the subsequent steps in our method (See Figure~\ref{fig:fig_flowchart}). 

Once the latent space is constructed, we apply clustering techniques to identify different modes within each class. These modes represent distinct subgroups or patterns within the class, reflecting the inherent diversity in the data. By clustering in the latent space, we can uncover all possible variations within a class, enabling us to better understand its internal structure. This step is crucial as it sets the stage for generating meaningful counterfactual explanations by highlighting the various directions in which data points can vary while still belonging to the same class.

After identifying the modes, we proceed to generate multiple counterfactual explanations by perturbing the data points along the directions identified through clustering. We use a linear classifier in conjunction with these clusters to determine the specific directions of perturbations that can lead to different outcomes. Notably, this approach does not require explicit concept labels; we only need class labels to guide the generation of counterfactuals. This makes our method both flexible and powerful, as it can be applied in situations where detailed annotations are unavailable, yet still provides valuable insights into the decision-making process of the model.

 The Diffusion Autoencoder effectively captures the most critical features of an image within its encodings, while a Diffusion Probabilistic Model (DPM) handles the remaining stochastic variations. These encodings have proven highly effective for classification, even with a simple linear classifier. By leveraging these encodings, we can generate multiple counterfactual explanations, which provide a more nuanced understanding of the internal variations within a class, thereby enhancing the interpretability of the model's decisions.

 Given a set of images $(I)$ and the corresponding labels $(Y)$ with $n$ distinct classes and within those classes there might be more than one mode ($M$), i.e. there might be various sub classes within a class as shown in Figure~\ref{fig:fig_melanoma}. Our method seek to generate counterfactual explanation images for each of these modes within a class. Firstly, the Diffusion Autoencoder ($Enc$) is trained such that $Enc(I_{i})=(z_{sem,i},x_{T,i})$. The encoding have two parts i) ($z_{sem}$) which captures the semantic information from the image and ii) $x_{T}$ which contains low level stochastic subcode. 

 Since most of the semantic information is captured by $z_{sem}$, we utilise that to find the various modes within a class. This is done with the help of clustering such that $f: Z_{sem} = \{C_{1},C_{2},...,C_{k}\}$. These cluster labels ($C$) can be used for generating the directions in which the $z_{sem}$ of an image is to be perturbed in order to generate counterfactual explanations. This can be done using a linear classifier on the semantic encoding with the cluster labels $g:Z_{sem} \rightarrow C$. Corresponding to each of these clusters ($C_{i}$) we get direction ($w_{i}$). Next, perturbations can be made to an encoding of an image ($z_{sem  i}$) such that $z'_{sem i}=z_{sem i} + \alpha * w_{j}$, where $w_{j}$ are the weights of linear classifier corresponding to cluster ($C_{j}$) and $\alpha$ is an adjustable parameter. Finally, counterfactual explanations corresponding to cluster $C_{j}$  can be generated such that $I'_{i,j}=Dec(z'_{sem,i},x_{T,i})$.
 %Given a classifier $C: X \rightarrow Y$ where $X$ are the set of encodings, and $Y$ is the set of labels, we wish to be able to generate counterfactual explanations, where the encoding $x_{i}$ is the perturbed to get $x_{i}'$ such that $C(x_{i})=y_{i}, C(x_{i}')=y_{j}$ and $\vert x_{i}-x_{i}'\vert <\epsilon$. Now an additional constraint is that each class label $Y_{j}$ has $k$ intra-class variations or modes $M_{jk}$ for membership, representing different feature combinations. We wish to generate counterfactuals that cover all possible modes i.e. we wish to generate all possible perturbations for an image to be translated from class $y_{i}$ to class $y_{j}$. %
%, if we are able to generate multiple counterfactual explanations using the diffusion autoencode

\subsection{Comparison between DISSECT and DifCluE}

% Dissect model is able to generate multiple counterfactual explanations that are distinct. The extent of the perturbations made to generate the counterfactual explanations, can be adjusted using the hyper parameter $\alpha$. The distinct counterfactual explanations are discovered in an unsupervised manner. DISSECT model relies on discovering multiple concepts which when perturbed leads to a classification change. It does not require labels for concepts, but just the class labels.

% Our method DifCluE on the other hand also aims to generate multiple distinct counterfactual explanations. But in our case, we first train an Encoder-Decoder that is based on diffusion model. We use the latent space from encoder part for clustering. This helps to identify all possible modes within a class. Once the modes are identified, they can be used to generate multiple counterfactual explanations. Thereafter, we perform clustering followed by a linear classifier to obtain the direction of perturbations. Importantly, this step also does not require concept labels. We just need class labels to generate counterfactual explanations.  

% One big difference between DISSECT and DifCluE is that the DISSECT approach would require retraining, if we need to increase the number of counterfactual factual explanation to be generated. Interestingly, our proposed approach i.e. DifCluE could easily change the number of counterfactual explanations, without requiring any retraining of the entire network. Therefore, if we look at the computational cost, DifClue is much better option than DISSECT approach.

Our \textit{DifCluE} approach is meticulously designed to generate a diverse set of counterfactual explanations, beginning with a Diffusion Autoencoder as its core. The process starts with training an Encoder-Decoder model grounded in diffusion processes, which excels at capturing the most salient features of input data within its latent space. This latent space acts as a compressed yet information-rich representation of each data point, preserving the essential characteristics needed for the subsequent analysis and generation of counterfactuals.

With the latent space established, we employ sophisticated clustering techniques to discern the various modes present within each class. These modes correspond to distinct subgroups or patterns within the class, encapsulating the inherent variability and complexity of the data. Clustering in the latent space allows us to uncover all possible variations that exist within a class, offering a deeper understanding of its internal dynamics. This step is pivotal, as it sets the foundation for generating meaningful and contextually relevant counterfactual explanations by revealing the different trajectories along which data points can be altered while maintaining their class membership.

Following the identification of these modes, we generate multiple counterfactual explanations by strategically perturbing data points in the directions indicated by the clustering results. To refine this process, we integrate a linear classifier with these clusters, which helps pinpoint the specific directions of perturbation that are most likely to produce alternative outcomes. A key advantage of our method is that it does not require detailed concept labels; class labels alone suffice to drive the generation of counterfactuals. This feature makes \textit{DifCluE} both adaptable and robust, enabling its application across various domains where comprehensive annotations are scarce, while still providing deep insights into the decision-making processes of the model.

A major distinction between \textit{DISSECT} and \textit{DifCluE} lies in their flexibility regarding the number of counterfactual explanations generated. DISSECT requires retraining the model if the number of desired counterfactuals increases, which adds to the computational burden. In contrast, \textit{DifCluE} allows for easy adjustment of the number of counterfactual explanations without necessitating any retraining of the network. This makes \textit{DifCluE} a more computationally efficient and scalable option compared to the \textit{DISSECT} approach.

%\subsection{DISSECT}
%DISSECT is a GAN based model which claims to be able to generate multiple counterfactual explanations, it is able to do so in an unsupervised manner using a distinctness loss function.
%
%For our experiments we train the DISSECT model also on the CelebA Dataset, where the Bang hairstyle and Blonde hair classes were deliberately assigned the same positive labels. Given this data we test if the model is able to discover two distinct ways of converting a negative image not belonging to that class to a positive image. Ideally, one of the counterfactual explanations would give the person in the image blonde hair while the other counterfactual explanation would give the person bangs.%

\subsection{Evaluation of Counterfactual Explanations}

In order to assess the quality of the counterfactual explanations generated by each of these models, we used various evaluation parameters to assess the following properties \cite{DBLP:journals/corr/abs-2105-15164}:

\begin{enumerate}

    \item \textbf{Realism}: The counterfactual explanations should lie on the data manifold, often referred to as realism. Frechet Inception Distance (FID) are based on how similar the two groups are in terms of statistics on computer vision features of the raw images calculated using the Inception v3 model used for image classification.\cite{heusel2018ganstrainedtimescaleupdate}
    
    \item \textbf{Substitutability}: The representation of a sample in terms of concepts should preserve relevant information. Substitutability measures an external classifier’s performance on real data when it is trained using only synthetic images.
    
    \item \textbf{Importance}: Explanations should produce the desired outcome as we make larger and larger perturbations. As we increase the parameter for the perturbation the classification probability for the target probability should also increase. 
    
    \item \textbf{Distinctness}: Another desirable quality for explanations is to represent inputs with non-overlapping concepts, often referred to as diversity. We want to access if the produced counterfactual explanations are distinct from one another or not. 

\end{enumerate}

\subsection{Evaluating the alignment of Counterfactual Explanations with actual classes}

Our aim here is to objectively evaluate the quality of the counterfactual explanations generated by the two different models, as well as evaluate how much  the counterfactual explanations correspond to the actual class that we mixed. In order to do so, we use a separate ResNet model trained to identify the 40 attributes of the CelebA dataset. The accuracy of the model in identifying the features is as shown in Table~\ref{table_resnet}.

\begin{table}[h]
\centering
%\resizebox{.95\columnwidth}{!}{
\begin{tabular}{|l|l|l|l|} \hline
     & Precision & Recall & F-1 score \\ \hline
    Blonde Hair &  0.803& 0.813 & 0.808 \\ \hline
    Bangs Hairstyle & 0.807 & 0.888 & 0.845 \\\hline
    Attractive & 0.837 & 0.858 & 0.851 \\\hline
    Eyeglasses & 0.891 & 0.912 & 0.902 \\\hline
    Young & 0.817 & 0.889 & 0.861 \\\hline
    Smiling & 0.927 & 0.968 & 0.941 \\\hline
    Mouth Open & 0.871 & 0.827 & 0.851 \\\hline
    Heavy Makeup & 0.891 & 0.912 & 0.902 \\\hline
    
\end{tabular}
\caption{ResNet Classification Performance}
\label{table_resnet}
\end{table}

The idea here is to take images that were labelled  negative (for example, neither blonde hair nor bangs hairstyle) and generate the two counterfactual explanations, further these images would be fed to this ResNet model and the prediction were recorded. Ideally, one of the counterfactual explanation should lead to blonde hair classification while the other should lead to bangs hairstyle. We note down what proportion of the counterfactual explanation images, undergo a class change to `Blonde Hair' or `Bangs Hairstyle'.  This helps us to understand if the counterfactual explanations align with the correct attributes or not.

\section{Experimental Results}

We first use the FFHQ dataset to train the Diffusion Autoencoder \cite{Karras_21}. Our aim here is to train on a large dataset. Thereafter, we use CelebA dataset to evaluate \textit{DifCluE}'s ability to generate multiple counterfactual explanations. CelebA dataset has labels for 40 attributes, including separate labels for Celebs with Bangs, and Celebs with blondes~\cite{liu2015faceattributes}. These labels are essential to evaluate the counterfactual explanations in a quantifiable manner. 

For our experiments, the Bang hairstyle and Blonde hair classes were deliberately assigned the same labels. The images which were labelled in this manner, were then fed to the autoencoder to get their latent representation. K-means clustering was applied on this latent representation to obtain the two clusters. Using the linear classifier, we get the direction for our perturbations and generate counterfactual explanations by perturbing the images along these directions.

We first assess the quality of the counterfactual explanations generated by each of these models. We used various evaluation parameters mentioned in methodology section to assess this. From Table \ref{table_realism}, we observe that the FID scores seem to suggest that our model out performs the DISSECT model by a clear margin. Therefore, we can say that in terms of realism, the \textit{DifCluE} model seems to be generating slightly better quality images.%Further table also show the accuracy, precision and recall for an external CNN model in identifying whether the image was a generated image or not. There we see that for both model the external classifier is not able to distinguish between generated and original images. Therefore, we can say that in terms of realism the DifCluE model seems to be generating slightly better quality images.

Table \ref{table_distinctness} seem to suggest that the distinctness between the two counterfactual explanations is closely matched. Distinctness between the two counterfactual explanations is measured by an external CNN classifier trained on classifying if the image belongs to counterfactual explanation 1 or 2. The accuracy and precision are better in case of \textit{DISSECT} model however the recalls are better in case of \textit{DifCluE}. However, the results are quite close to each other suggesting the two models perform similar accuracy levels. Table \ref{table_substitutability} compare the substitutability for the models. Substitutability measures an external classifier’s performance on real data, when it is trained using only synthetic images. Here, it can be seen that the performance of \textit{DISSECT} and \textit{DifCluE} model is comparable.

%The idea there is both DISSECT and DifCluE allow us to make perturbation which is controlled by hyper-parameter ($\alpha$).

Importance property is compared in Table \ref{table_importance}.  The idea here is that, as we make bigger and bigger perturbations, the classification score of the generated images must also increase for that particular class along which perturbations are made. This suggests that the results are quite similar. KL score and MSE score are both better in case of \textit{DISSECT} model; however R-value and the correlation coefficient are better for \textit{DifCluE} model. We can observe from Tables \ref{table_realism},\ref{table_distinctness},\ref{table_substitutability},\ref{table_importance} that for most of the evaluation parameters, DifCluE and the DISSECT model perform quite comparably.

\begin{table}[h]
\centering
%\resizebox{.95\columnwidth}{!}{
    \begin{tabular}{|l|l|} \hline
          &$\downarrow$ FID Score\\ \hline
        DISSECT &  11.0 \\ \hline
        DifCluE & \textbf{9.26} \\\hline
        
    \end{tabular}
    \caption{Realism Comparison}
\label{table_realism}
\end{table}

\begin{table}[h]
\centering
%\resizebox{.95\columnwidth}{!}{
\begin{tabular}{|p{13mm}|p{9mm}|p{10mm}|p{10mm}|p{10mm}|p{10mm}|} \hline
    &$\uparrow$ Acc&  \shortstack{$\uparrow$ Prec\\(micro)}&\shortstack{$\uparrow$Prec\\(macro)} & \shortstack{$\uparrow$ Rec\\(micro)} & \shortstack{$\uparrow$ Rec\\(macro)} \\ \hline
    DISSECT &  \textbf{0.95} & \textbf{0.98} & \textbf{0.98} & 0.96 & 0.961 \\ \hline
    DifCluE & 0.93 & 0.953 & 0.951 & \textbf{0.963} & \textbf{0.964}\\\hline
    
\end{tabular}
\caption{Distinctness Comparison}
\label{table_distinctness}
\end{table}

\begin{table}[h]
\centering
%\resizebox{.95\columnwidth}{!}{
\begin{tabular}{|l|l|l|l|} \hline
       &$\uparrow$ Accuracy&$\uparrow$ Precision&$\uparrow$ Recall\\ \hline
    DISSECT &  \textbf{91.9} & \textbf{96.9} & \textbf{87.6}\\ \hline
    DifCluE & 88.1 & 90.4 & 86.5\\\hline
    
\end{tabular}
\caption{Substitutability Comparison}
\label{table_substitutability}
\end{table}

\begin{table}[h]
\centering
%\resizebox{.95\columnwidth}{!}{
\begin{tabular}{|l|l|l|l|l|} \hline
       &$\uparrow$ R& $\uparrow \rho$& $\downarrow$ KL&$\downarrow$ MSE\\ \hline
    DISSECT &  0.84 & 0.88 & \textbf{0.19} & \textbf{0.047}\\ \hline
    DifCluE & \textbf{0.85} & \textbf{0.90} & 0.22 & 0.091\\ \hline %\hline
\end{tabular}
\caption{Importance Comparison}
\label{table_importance}
\end{table}

\begin{table*}[h]
\centering
%\resizebox{.95\columnwidth}{!}{
\begin{tabular}{|p{2.5cm}|p{2.5cm}|p{2.5cm}|p{2.5cm}|} \hline
    \multicolumn{2}{|c|}{\textbf{DifCluE}}&\multicolumn{2}{c|}{\textbf{DISSECT}}\\ \hline
     Class & Percent Conversion & Class & Percent Conversion \\ \hline
    Blonde Hair &  \textbf{0.88} & Blonde Hair &  0.51\\ \hline
    Bangs Hairstyle & \textbf{0.78} & Bangs Hairstyle & 0.33\\\hline
    
    \multicolumn{2}{|c|}{DifCluE Sample Perturbations}&\multicolumn{2}{c|}{DISSECT Sample Perturbations}\\ \hline
    Cluster 1 & Cluster 2 & Cluster 1 & Cluster 2\\ \hline
    \hspace{-0.15cm}\parbox[c]{1em}{ \includegraphics[width=1.1in]{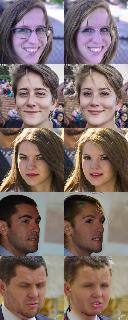}} & \hspace{-0.15cm}\parbox[c]{1em}{ \includegraphics[width=1.1in]{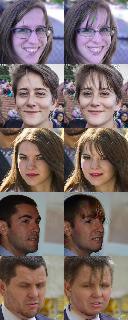}}&
    \hspace{-0.15cm}\parbox[c]{1em}{ \includegraphics[width=1.1in]{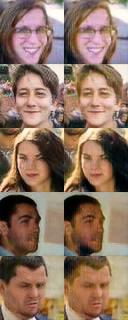}} & 
    \hspace{-0.15cm}\parbox[c]{1em}{ \includegraphics[width=1.1in]{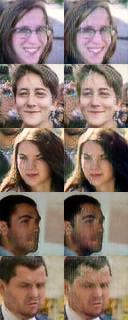}}\\\hline
\end{tabular}
\caption{Alignment of counterfactual explanations with the Actual classes (Blond Hair and Bangs) that were mixed.}
\label{table_DiffAE_res1}
\end{table*}

\begin{table*}[h]
\centering
%\resizebox{.95\columnwidth}{!}{
\begin{tabular}{|p{2.5cm}|p{2.5cm}|p{2.5cm}|p{2.5cm}|} \hline
    \multicolumn{2}{|c|}{\textbf{DifCluE}}&\multicolumn{2}{c|}{\textbf{DISSECT}}\\ \hline
     Class & Percent Conversion & Class & Percent Conversion \\ \hline
     Young &  \textbf{0.85} & Young & 0.61 \\ \hline
     Smiling & \textbf{0.87} & Smiling & 0.51\\\hline
    \multicolumn{2}{|c|}{DifCluE Sample Perturbations} & \multicolumn{2}{c|}{DISSECT Sample Perturbations}\\ \hline
    Cluster 1 & Cluster 2 & Cluster 1 & Cluster 2\\ \hline
    \hspace{-0.15cm}\parbox[c]{1em}{ \includegraphics[width=1.1in]{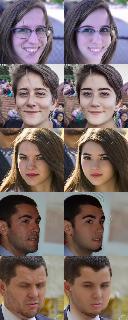}} & \hspace{-0.15cm}\parbox[c]{1em}{ \includegraphics[width=1.1in]{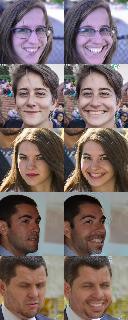}}&
    \hspace{-0.15cm}\parbox[c]{1em}{ \includegraphics[width=1.1in]{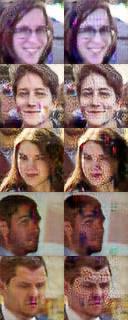}} & \hspace{-0.15cm}\parbox[c]{1em}{ \includegraphics[width=1.1in]{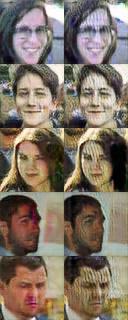}}\\\hline
    
\end{tabular}
\caption{Alignment of counterfactual explanations with the Actual classes (Young and Smiling) that were mixed.}
\label{table_DiffAE_res2}
\end{table*}

\begin{table*}[h]
\centering
%\resizebox{.95\columnwidth}{!}{
\begin{tabular}{|p{2.5cm}|p{2.5cm}|p{2.5cm}|p{2.5cm}|} \hline
    \multicolumn{2}{|c|}{\textbf{DifCluE}}&\multicolumn{2}{c|}{\textbf{DISSECT}}\\ \hline
     Class & Percent Conversion & Class & Percent Conversion\\ \hline
    Attractive &  \textbf{0.87} & Attractive &  0.65 \\ \hline
    Eyeglasses & \textbf{0.75} & Eyeglasses &  0.59\\\hline
    % \multicolumn{2}{c}{DISSECT}\\ \hline
    % Attractive &  0.65\\ \hline
    % Eyeglasses &  0.59\\\hline
    \multicolumn{2}{|c|}{DifCluE Sample Perturbations}&\multicolumn{2}{c|}{DISSECT Sample Perturbations}\\ \hline
    Cluster 1 & Cluster 2 & Cluster 1 & Cluster 2\\ \hline
    \hspace{-0.15cm}\parbox[c]{1em}{ \includegraphics[width=1.1in]{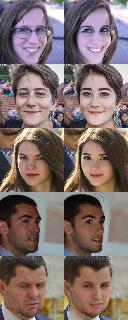}} & \hspace{-0.15cm}\parbox[c]{1em}{ \includegraphics[width=1.1in]{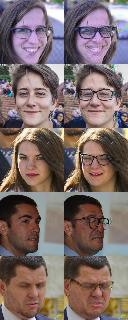}}&
    \hspace{-0.15cm}\parbox[c]{1em}{ \includegraphics[width=1.1in]{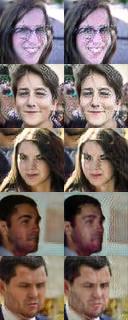}} & \hspace{-0.15cm}\parbox[c]{1em}{ \includegraphics[width=1.1in]{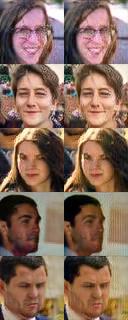}}\\\hline
\end{tabular}
\caption{Alignment of counterfactual explanations with the Actual classes (Attractive and Eyeglasses) that were mixed.}
\label{table_DiffAE_res3}
\end{table*}
% \vspace{-em}

\begin{table*}[h]
\centering
%\resizebox{.95\columnwidth}{!}{
\begin{tabular}{|p{2.5cm}|p{2.5cm}|p{2.5cm}|p{2.5cm}|} \hline
    \multicolumn{2}{|c|}{\textbf{DifCluE}}&\multicolumn{2}{c|}{\textbf{DISSECT}}\\ \hline
     Class & Percent Conversion & Class & Percent Conversion  \\ \hline
    Mouth Slightly Open &  \textbf{0.73} & Mouth Slightly Open &   0.48\\ \hline
    Heavy Makeup & \textbf{0.81} & Heavy Makeup &  0.61\\\hline
    % \multicolumn{2}{c}{DISSECT}\\ \hline
    % \\ \hline
    % \\ \hline
    % \\\hline
    
    \multicolumn{2}{|c|}{DifCluE Sample Perturbations}& \multicolumn{2}{c|}{DISSECT Sample Perturbations}\\ \hline
    Cluster 1 & Cluster 2&  Cluster 1 & Cluster 2\\ \hline
    \hspace{-0.15cm}\parbox[c]{1em}{ \includegraphics[width=1.1in]{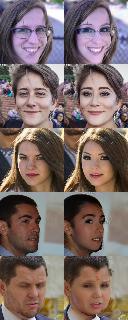}} & \hspace{-0.15cm}\parbox[c]{1em}{ \includegraphics[width=1.1in]{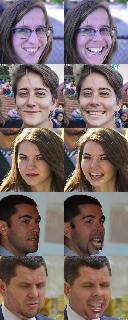}}&
    \hspace{-0.15cm}\parbox[c]{1em}{ \includegraphics[width=1.1in]{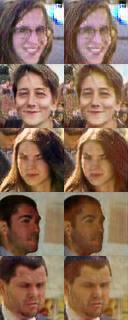}} & \hspace{-0.15cm}\parbox[c]{1em}{ \includegraphics[width=1.1in]{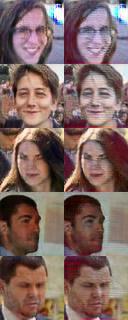}}\\ \hline
    
    % \multicolumn{2}{c}{DiffAE Sample Perturbations}\\ \hline
    % Cluster 1 & Cluster 2\\ \hline
    % \parbox[c]{1em}{ \includegraphics[width=1in]{Submission-2024/LaTeX/clustering_res/open_mouth.png}} & \parbox[c]{1em}{ \includegraphics[width=1in]{Submission-2024/LaTeX/clustering_res/heavy_makeup.png}}\\\hline
\end{tabular}
\caption{Alignment of counterfactual explanations with the Actual classes (Mouth Slightly Open and Heavy Makeup) that were mixed.}
\label{table_DiffAE_res4}
\end{table*}

Next, we also want to evaluate the disentanglement of the generated counterfactual explanations. In order to do so, firstly we train an external classifier capable of classifying the images into the 40 categories labelled in the dataset. Once we have this classifier, the generated counterfactual explanations are fed this network. If we have a negative sample, we could generate two distinct counterfactual explanations and these counterfactual explanations should be aligned with one of the classes, since the positive samples in our experiments were obtained by mixing two classes. 

Therefore with the help of our current pipeline, we should be able to disentangle the two classes that we had mixed in this experiment. Hence, it is expected that the two counterfactual explanations should be aligned with the two constituent classes. For example, if we trained a Diffusion Autoencoder on data where both `Bangs' and `Blondes' were classified in the same class and the rest of the samples were labeled as `negative', so the generated counter factual explanations should be images that are `Bangs' and `Blondes'. Hence these images should be classified by the external classifier as such. We have here recorded the percentage of the counterfactual explanations that got the desired results. The results for this example are summarized in Table\ref{table_DiffAE_res1}.

We conducted this experiment with 4 different pairs of classes being mixed (Tables: \ref{table_DiffAE_res1},\ref{table_DiffAE_res2},\ref{table_DiffAE_res3},\ref{table_DiffAE_res4}). From these set of experiments we observe that in each case, \textit{DifCluE} outperformed the \textit{DISSECT} model. The counterfactual images that \textit{DISSECT} generated were only being classified as either of the two constituent classes, 53 per cent of the time as compared to 83 percent of the times in case of \textit{DifCluE}. Therefore, it is observed that the \textit{DifCluE} approach performed much better in generating disentangled and distinct counterfactual explanations.

\section{Conclusion}

% We present a methodology to generate counterfactual exemplars for each mode (or sub-class) in a given target class. Our exemplars are based on perturbations at the conceptual feature level, rather than at the low-level pixel level. Our method clusters the input encodings in the latent space to gather the distinctness between various modes in a class. This information is used in generating meaningful counterfactuals for each mode by determining the kind of perturbations to do in order to generate a counterfactual for a mode. Since the perturbations are carried out at the semantic level, the resulting explainability of the counterfactual exemplars is much better and therefore it can be easily understood by the end-users. This also helps in understanding the conceptual feature level changes that influence a classification into a target class leading to improve trust in the underlying model. In our future work, we would like to try this in other domains such as time-series applications.

We have introduced a methodology for generating counterfactual exemplars for each mode (or sub-class) within a given target class, focusing on perturbations at the conceptual feature level rather than the low-level pixel level. By clustering input encodings in the latent space, our method effectively captures the distinct characteristics of various modes within a class. This information is then leveraged to generate meaningful counterfactuals by determining the specific perturbations needed to produce a counterfactual for each mode. Since these perturbations are applied at the semantic level, the resulting counterfactual exemplars offer enhanced explainability, making them more intuitive and accessible to end-users. This approach not only clarifies the conceptual feature-level changes that influence classification decisions but also enhances trust in the underlying model. In future work, we plan to extend this approach to other domains, such as time-series applications, to further explore its versatility and impact.

\bibliographystyle{ACM-Reference-Format}
\bibliography{DifCluE_CIKM_24}


\begin{thebibliography}{29}

%%% the Web address.

\ifx \showCODEN    \undefined \def \showCODEN     #1{\unskip}     \fi
\ifx \showDOI      \undefined \def \showDOI       #1{#1}\fi
\ifx \showISBNx    \undefined \def \showISBNx     #1{\unskip}     \fi
\ifx \showISBNxiii \undefined \def \showISBNxiii  #1{\unskip}     \fi
\ifx \showISSN     \undefined \def \showISSN      #1{\unskip}     \fi
\ifx \showLCCN     \undefined \def \showLCCN      #1{\unskip}     \fi
\ifx \shownote     \undefined \def \shownote      #1{#1}          \fi
\ifx \showarticletitle \undefined \def \showarticletitle #1{#1}   \fi
\ifx \showURL      \undefined \def \showURL       {\relax}        \fi
% invisible to TeX
\providecommand\bibfield[2]{#2}
\providecommand\bibinfo[2]{#2}
\providecommand\natexlab[1]{#1}
\providecommand\showeprint[2][]{arXiv:#2}

\bibitem[Avrahami et~al\mbox{.}(2021)]%
        {DBLP:journals/corr/abs-2111-14818}
\bibfield{author}{\bibinfo{person}{Omri Avrahami}, \bibinfo{person}{Dani
  Lischinski}, {and} \bibinfo{person}{Ohad Fried}.}
  \bibinfo{year}{2021}\natexlab{}.
\newblock \showarticletitle{Blended Diffusion for Text-driven Editing of
  Natural Images}.
\newblock \bibinfo{journal}{\emph{CoRR}}  \bibinfo{volume}{abs/2111.14818}
  (\bibinfo{year}{2021}).
\newblock
\showeprint[arXiv]{2111.14818}
\urldef\tempurl%
\url{https://arxiv.org/abs/2111.14818}
\showURL{%
\tempurl}


\bibitem[Barberan et~al\mbox{.}(2022)]%
        {barberan2022neuroviewrnn}
\bibfield{author}{\bibinfo{person}{CJ Barberan}, \bibinfo{person}{Sina
  Alemohammad}, \bibinfo{person}{Naiming Liu}, \bibinfo{person}{Randall
  Balestriero}, {and} \bibinfo{person}{Richard~G. Baraniuk}.}
  \bibinfo{year}{2022}\natexlab{}.
\newblock \bibinfo{title}{NeuroView-RNN: It's About Time}.
\newblock
\newblock
\showeprint[arxiv]{2202.11811}~[cs.LG]


\bibitem[Bengio et~al\mbox{.}(2012)]%
        {DBLP:journals/corr/abs-1206-5538}
\bibfield{author}{\bibinfo{person}{Yoshua Bengio}, \bibinfo{person}{Aaron~C.
  Courville}, {and} \bibinfo{person}{Pascal Vincent}.}
  \bibinfo{year}{2012}\natexlab{}.
\newblock \showarticletitle{Unsupervised Feature Learning and Deep Learning:
  {A} Review and New Perspectives}.
\newblock \bibinfo{journal}{\emph{CoRR}}  \bibinfo{volume}{abs/1206.5538}
  (\bibinfo{year}{2012}).
\newblock
\showeprint[arXiv]{1206.5538}
\urldef\tempurl%
\url{http://arxiv.org/abs/1206.5538}
\showURL{%
\tempurl}


\bibitem[Carlevaro et~al\mbox{.}(2023)]%
        {Carlevaro2023}
\bibfield{author}{\bibinfo{person}{Alberto Carlevaro}, \bibinfo{person}{Marta
  Lenatti}, \bibinfo{person}{Alessia Paglialonga}, {and}
  \bibinfo{person}{Maurizio Mongelli}.} \bibinfo{year}{2023}\natexlab{}.
\newblock \showarticletitle{{Multi-Class Counterfactual Explanations using
  Support Vector Data Description}}.
\newblock  (\bibinfo{date}{3} \bibinfo{year}{2023}).
\newblock
\urldef\tempurl%
\url{https://doi.org/10.36227/techrxiv.22221007.v1}
\showDOI{\tempurl}


\bibitem[Chen et~al\mbox{.}(2018)]%
        {chen2018isolating}
\bibfield{author}{\bibinfo{person}{Ricky~TQ Chen}, \bibinfo{person}{Xuechen
  Li}, \bibinfo{person}{Roger~B Grosse}, {and} \bibinfo{person}{David~K
  Duvenaud}.} \bibinfo{year}{2018}\natexlab{}.
\newblock \showarticletitle{Isolating sources of disentanglement in variational
  autoencoders}.
\newblock \bibinfo{journal}{\emph{Advances in neural information processing
  systems}}  \bibinfo{volume}{31} (\bibinfo{year}{2018}).
\newblock


\bibitem[Choi et~al\mbox{.}(2021)]%
        {DBLP:journals/corr/abs-2108-02938}
\bibfield{author}{\bibinfo{person}{Jooyoung Choi}, \bibinfo{person}{Sungwon
  Kim}, \bibinfo{person}{Yonghyun Jeong}, \bibinfo{person}{Youngjune Gwon},
  {and} \bibinfo{person}{Sungroh Yoon}.} \bibinfo{year}{2021}\natexlab{}.
\newblock \showarticletitle{{ILVR:} Conditioning Method for Denoising Diffusion
  Probabilistic Models}.
\newblock \bibinfo{journal}{\emph{CoRR}}  \bibinfo{volume}{abs/2108.02938}
  (\bibinfo{year}{2021}).
\newblock
\showeprint[arXiv]{2108.02938}
\urldef\tempurl%
\url{https://arxiv.org/abs/2108.02938}
\showURL{%
\tempurl}


\bibitem[Dhariwal and Nichol(2021)]%
        {DBLP:journals/corr/abs-2105-05233}
\bibfield{author}{\bibinfo{person}{Prafulla Dhariwal} {and}
  \bibinfo{person}{Alex Nichol}.} \bibinfo{year}{2021}\natexlab{}.
\newblock \showarticletitle{Diffusion Models Beat GANs on Image Synthesis}.
\newblock \bibinfo{journal}{\emph{CoRR}}  \bibinfo{volume}{abs/2105.05233}
  (\bibinfo{year}{2021}).
\newblock
\showeprint[arXiv]{2105.05233}
\urldef\tempurl%
\url{https://arxiv.org/abs/2105.05233}
\showURL{%
\tempurl}


\bibitem[Downs et~al\mbox{.}(2020)]%
        {downs2020cruds}
\bibfield{author}{\bibinfo{person}{Michael Downs}, \bibinfo{person}{Jonathan~L
  Chu}, \bibinfo{person}{Yaniv Yacoby}, \bibinfo{person}{Finale Doshi-Velez},
  {and} \bibinfo{person}{Weiwei Pan}.} \bibinfo{year}{2020}\natexlab{}.
\newblock \showarticletitle{Cruds: Counterfactual recourse using disentangled
  subspaces}.
\newblock \bibinfo{journal}{\emph{ICML WHI}}  \bibinfo{volume}{2020}
  (\bibinfo{year}{2020}), \bibinfo{pages}{1--23}.
\newblock


\bibitem[Ghandeharioun et~al\mbox{.}(2021)]%
        {DBLP:journals/corr/abs-2105-15164}
\bibfield{author}{\bibinfo{person}{Asma Ghandeharioun}, \bibinfo{person}{Been
  Kim}, \bibinfo{person}{Chun{-}Liang Li}, \bibinfo{person}{Brendan Jou},
  \bibinfo{person}{Brian Eoff}, {and} \bibinfo{person}{Rosalind~W. Picard}.}
  \bibinfo{year}{2021}\natexlab{}.
\newblock \showarticletitle{{DISSECT:} Disentangled Simultaneous Explanations
  via Concept Traversals}.
\newblock \bibinfo{journal}{\emph{CoRR}}  \bibinfo{volume}{abs/2105.15164}
  (\bibinfo{year}{2021}).
\newblock
\showeprint[arXiv]{2105.15164}
\urldef\tempurl%
\url{https://arxiv.org/abs/2105.15164}
\showURL{%
\tempurl}


\bibitem[Heusel et~al\mbox{.}(2018)]%
        {heusel2018ganstrainedtimescaleupdate}
\bibfield{author}{\bibinfo{person}{Martin Heusel}, \bibinfo{person}{Hubert
  Ramsauer}, \bibinfo{person}{Thomas Unterthiner}, \bibinfo{person}{Bernhard
  Nessler}, {and} \bibinfo{person}{Sepp Hochreiter}.}
  \bibinfo{year}{2018}\natexlab{}.
\newblock \bibinfo{title}{GANs Trained by a Two Time-Scale Update Rule Converge
  to a Local Nash Equilibrium}.
\newblock
\newblock
\showeprint[arxiv]{1706.08500}~[cs.LG]
\urldef\tempurl%
\url{https://arxiv.org/abs/1706.08500}
\showURL{%
\tempurl}


\bibitem[Heusel et~al\mbox{.}(2017)]%
        {DBLP:journals/corr/HeuselRUNKH17}
\bibfield{author}{\bibinfo{person}{Martin Heusel}, \bibinfo{person}{Hubert
  Ramsauer}, \bibinfo{person}{Thomas Unterthiner}, \bibinfo{person}{Bernhard
  Nessler}, \bibinfo{person}{G{\"{u}}nter Klambauer}, {and}
  \bibinfo{person}{Sepp Hochreiter}.} \bibinfo{year}{2017}\natexlab{}.
\newblock \showarticletitle{GANs Trained by a Two Time-Scale Update Rule
  Converge to a Nash Equilibrium}.
\newblock \bibinfo{journal}{\emph{CoRR}}  \bibinfo{volume}{abs/1706.08500}
  (\bibinfo{year}{2017}).
\newblock
\showeprint[arXiv]{1706.08500}
\urldef\tempurl%
\url{http://arxiv.org/abs/1706.08500}
\showURL{%
\tempurl}


\bibitem[Higgins et~al\mbox{.}(2016)]%
        {higgins2016beta}
\bibfield{author}{\bibinfo{person}{Irina Higgins}, \bibinfo{person}{Loic
  Matthey}, \bibinfo{person}{Arka Pal}, \bibinfo{person}{Christopher Burgess},
  \bibinfo{person}{Xavier Glorot}, \bibinfo{person}{Matthew Botvinick},
  \bibinfo{person}{Shakir Mohamed}, {and} \bibinfo{person}{Alexander
  Lerchner}.} \bibinfo{year}{2016}\natexlab{}.
\newblock \showarticletitle{beta-vae: Learning basic visual concepts with a
  constrained variational framework}. In
  \bibinfo{booktitle}{\emph{International conference on learning
  representations}}.
\newblock


\bibitem[Ho et~al\mbox{.}(2020)]%
        {DBLP:journals/corr/abs-2006-11239}
\bibfield{author}{\bibinfo{person}{Jonathan Ho}, \bibinfo{person}{Ajay Jain},
  {and} \bibinfo{person}{Pieter Abbeel}.} \bibinfo{year}{2020}\natexlab{}.
\newblock \showarticletitle{Denoising Diffusion Probabilistic Models}.
\newblock \bibinfo{journal}{\emph{CoRR}}  \bibinfo{volume}{abs/2006.11239}
  (\bibinfo{year}{2020}).
\newblock
\showeprint[arXiv]{2006.11239}
\urldef\tempurl%
\url{https://arxiv.org/abs/2006.11239}
\showURL{%
\tempurl}


\bibitem[Jeanneret et~al\mbox{.}(2022)]%
        {jeanneret2022diffusion}
\bibfield{author}{\bibinfo{person}{Guillaume Jeanneret}, \bibinfo{person}{Loïc
  Simon}, {and} \bibinfo{person}{Frédéric Jurie}.}
  \bibinfo{year}{2022}\natexlab{}.
\newblock \bibinfo{title}{Diffusion Models for Counterfactual Explanations}.
\newblock
\newblock
\showeprint[arxiv]{2203.15636}~[cs.CV]


\bibitem[Kanehira et~al\mbox{.}(2019)]%
        {kanehira2019multimodal}
\bibfield{author}{\bibinfo{person}{Atsushi Kanehira}, \bibinfo{person}{Kentaro
  Takemoto}, \bibinfo{person}{Sho Inayoshi}, {and} \bibinfo{person}{Tatsuya
  Harada}.} \bibinfo{year}{2019}\natexlab{}.
\newblock \bibinfo{title}{Multimodal Explanations by Predicting
  Counterfactuality in Videos}.
\newblock
\newblock
\showeprint[arxiv]{1812.01263}~[cs.CV]


\bibitem[Karras et~al\mbox{.}(2021)]%
        {Karras_21}
\bibfield{author}{\bibinfo{person}{Tero Karras}, \bibinfo{person}{Samuli
  Laine}, {and} \bibinfo{person}{Timo Aila}.} \bibinfo{year}{2021}\natexlab{}.
\newblock \showarticletitle{A Style-Based Generator Architecture for Generative
  Adversarial Networks}.
\newblock \bibinfo{journal}{\emph{IEEE Trans. Pattern Anal. Mach. Intell.}}
  \bibinfo{volume}{43}, \bibinfo{number}{12} (\bibinfo{date}{dec}
  \bibinfo{year}{2021}), \bibinfo{pages}{4217–4228}.
\newblock
\showISSN{0162-8828}
\urldef\tempurl%
\url{https://doi.org/10.1109/TPAMI.2020.2970919}
\showDOI{\tempurl}


\bibitem[Kim and Mnih(2018)]%
        {kim2018disentangling}
\bibfield{author}{\bibinfo{person}{Hyunjik Kim} {and} \bibinfo{person}{Andriy
  Mnih}.} \bibinfo{year}{2018}\natexlab{}.
\newblock \showarticletitle{Disentangling by factorising}. In
  \bibinfo{booktitle}{\emph{International Conference on Machine Learning}}.
  PMLR, \bibinfo{pages}{2649--2658}.
\newblock


\bibitem[Kwon et~al\mbox{.}(2023)]%
        {kwon2023diffusion}
\bibfield{author}{\bibinfo{person}{Mingi Kwon}, \bibinfo{person}{Jaeseok
  Jeong}, {and} \bibinfo{person}{Youngjung Uh}.}
  \bibinfo{year}{2023}\natexlab{}.
\newblock \bibinfo{title}{Diffusion Models already have a Semantic Latent
  Space}.
\newblock
\newblock
\showeprint[arxiv]{2210.10960}~[cs.CV]


\bibitem[Liu et~al\mbox{.}(2021)]%
        {DBLP:journals/corr/abs-2112-05744}
\bibfield{author}{\bibinfo{person}{Xihui Liu}, \bibinfo{person}{Dong~Huk Park},
  \bibinfo{person}{Samaneh Azadi}, \bibinfo{person}{Gong Zhang},
  \bibinfo{person}{Arman Chopikyan}, \bibinfo{person}{Yuxiao Hu},
  \bibinfo{person}{Humphrey Shi}, \bibinfo{person}{Anna Rohrbach}, {and}
  \bibinfo{person}{Trevor Darrell}.} \bibinfo{year}{2021}\natexlab{}.
\newblock \showarticletitle{More Control for Free! Image Synthesis with
  Semantic Diffusion Guidance}.
\newblock \bibinfo{journal}{\emph{CoRR}}  \bibinfo{volume}{abs/2112.05744}
  (\bibinfo{year}{2021}).
\newblock
\showeprint[arXiv]{2112.05744}
\urldef\tempurl%
\url{https://arxiv.org/abs/2112.05744}
\showURL{%
\tempurl}


\bibitem[Liu et~al\mbox{.}(2015)]%
        {liu2015faceattributes}
\bibfield{author}{\bibinfo{person}{Ziwei Liu}, \bibinfo{person}{Ping Luo},
  \bibinfo{person}{Xiaogang Wang}, {and} \bibinfo{person}{Xiaoou Tang}.}
  \bibinfo{year}{2015}\natexlab{}.
\newblock \showarticletitle{Deep Learning Face Attributes in the Wild}. In
  \bibinfo{booktitle}{\emph{Proceedings of International Conference on Computer
  Vision (ICCV)}}.
\newblock


\bibitem[Locatello et~al\mbox{.}(2018)]%
        {DBLP:journals/corr/abs-1811-12359}
\bibfield{author}{\bibinfo{person}{Francesco Locatello},
  \bibinfo{person}{Stefan Bauer}, \bibinfo{person}{Mario Lucic},
  \bibinfo{person}{Sylvain Gelly}, \bibinfo{person}{Bernhard Sch{\"{o}}lkopf},
  {and} \bibinfo{person}{Olivier Bachem}.} \bibinfo{year}{2018}\natexlab{}.
\newblock \showarticletitle{Challenging Common Assumptions in the Unsupervised
  Learning of Disentangled Representations}.
\newblock \bibinfo{journal}{\emph{CoRR}}  \bibinfo{volume}{abs/1811.12359}
  (\bibinfo{year}{2018}).
\newblock
\showeprint[arXiv]{1811.12359}
\urldef\tempurl%
\url{http://arxiv.org/abs/1811.12359}
\showURL{%
\tempurl}


\bibitem[Lugmayr et~al\mbox{.}(2022)]%
        {DBLP:journals/corr/abs-2201-09865}
\bibfield{author}{\bibinfo{person}{Andreas Lugmayr}, \bibinfo{person}{Martin
  Danelljan}, \bibinfo{person}{Andr{\'{e}}s Romero}, \bibinfo{person}{Fisher
  Yu}, \bibinfo{person}{Radu Timofte}, {and} \bibinfo{person}{Luc~Van Gool}.}
  \bibinfo{year}{2022}\natexlab{}.
\newblock \showarticletitle{RePaint: Inpainting using Denoising Diffusion
  Probabilistic Models}.
\newblock \bibinfo{journal}{\emph{CoRR}}  \bibinfo{volume}{abs/2201.09865}
  (\bibinfo{year}{2022}).
\newblock
\showeprint[arXiv]{2201.09865}
\urldef\tempurl%
\url{https://arxiv.org/abs/2201.09865}
\showURL{%
\tempurl}


\bibitem[Meng et~al\mbox{.}(2021)]%
        {DBLP:journals/corr/abs-2108-01073}
\bibfield{author}{\bibinfo{person}{Chenlin Meng}, \bibinfo{person}{Yang Song},
  \bibinfo{person}{Jiaming Song}, \bibinfo{person}{Jiajun Wu},
  \bibinfo{person}{Jun{-}Yan Zhu}, {and} \bibinfo{person}{Stefano Ermon}.}
  \bibinfo{year}{2021}\natexlab{}.
\newblock \showarticletitle{SDEdit: Image Synthesis and Editing with Stochastic
  Differential Equations}.
\newblock \bibinfo{journal}{\emph{CoRR}}  \bibinfo{volume}{abs/2108.01073}
  (\bibinfo{year}{2021}).
\newblock
\showeprint[arXiv]{2108.01073}
\urldef\tempurl%
\url{https://arxiv.org/abs/2108.01073}
\showURL{%
\tempurl}


\bibitem[Nemirovsky et~al\mbox{.}(2020)]%
        {DBLP:journals/corr/abs-2009-05199}
\bibfield{author}{\bibinfo{person}{Daniel Nemirovsky}, \bibinfo{person}{Nicolas
  Thiebaut}, \bibinfo{person}{Ye Xu}, {and} \bibinfo{person}{Abhishek Gupta}.}
  \bibinfo{year}{2020}\natexlab{}.
\newblock \showarticletitle{CounteRGAN: Generating Realistic Counterfactuals
  with Residual Generative Adversarial Nets}.
\newblock \bibinfo{journal}{\emph{CoRR}}  \bibinfo{volume}{abs/2009.05199}
  (\bibinfo{year}{2020}).
\newblock
\showeprint[arXiv]{2009.05199}
\urldef\tempurl%
\url{https://arxiv.org/abs/2009.05199}
\showURL{%
\tempurl}


\bibitem[Preechakul et~al\mbox{.}(2022)]%
        {preechakul2022diffusion}
\bibfield{author}{\bibinfo{person}{Konpat Preechakul},
  \bibinfo{person}{Nattanat Chatthee}, \bibinfo{person}{Suttisak Wizadwongsa},
  {and} \bibinfo{person}{Supasorn Suwajanakorn}.}
  \bibinfo{year}{2022}\natexlab{}.
\newblock \bibinfo{title}{Diffusion Autoencoders: Toward a Meaningful and
  Decodable Representation}.
\newblock
\newblock
\showeprint[arxiv]{2111.15640}~[cs.CV]


\bibitem[Sohl{-}Dickstein et~al\mbox{.}(2015)]%
        {DBLP:journals/corr/Sohl-DicksteinW15}
\bibfield{author}{\bibinfo{person}{Jascha Sohl{-}Dickstein},
  \bibinfo{person}{Eric~A. Weiss}, \bibinfo{person}{Niru Maheswaranathan},
  {and} \bibinfo{person}{Surya Ganguli}.} \bibinfo{year}{2015}\natexlab{}.
\newblock \showarticletitle{Deep Unsupervised Learning using Nonequilibrium
  Thermodynamics}.
\newblock \bibinfo{journal}{\emph{CoRR}}  \bibinfo{volume}{abs/1503.03585}
  (\bibinfo{year}{2015}).
\newblock
\showeprint[arXiv]{1503.03585}
\urldef\tempurl%
\url{http://arxiv.org/abs/1503.03585}
\showURL{%
\tempurl}


\bibitem[Song and Ermon(2019)]%
        {DBLP:journals/corr/abs-1907-05600}
\bibfield{author}{\bibinfo{person}{Yang Song} {and} \bibinfo{person}{Stefano
  Ermon}.} \bibinfo{year}{2019}\natexlab{}.
\newblock \showarticletitle{Generative Modeling by Estimating Gradients of the
  Data Distribution}.
\newblock \bibinfo{journal}{\emph{CoRR}}  \bibinfo{volume}{abs/1907.05600}
  (\bibinfo{year}{2019}).
\newblock
\showeprint[arXiv]{1907.05600}
\urldef\tempurl%
\url{http://arxiv.org/abs/1907.05600}
\showURL{%
\tempurl}


\bibitem[Song et~al\mbox{.}(2020)]%
        {DBLP:journals/corr/abs-2011-13456}
\bibfield{author}{\bibinfo{person}{Yang Song}, \bibinfo{person}{Jascha
  Sohl{-}Dickstein}, \bibinfo{person}{Diederik~P. Kingma},
  \bibinfo{person}{Abhishek Kumar}, \bibinfo{person}{Stefano Ermon}, {and}
  \bibinfo{person}{Ben Poole}.} \bibinfo{year}{2020}\natexlab{}.
\newblock \showarticletitle{Score-Based Generative Modeling through Stochastic
  Differential Equations}.
\newblock \bibinfo{journal}{\emph{CoRR}}  \bibinfo{volume}{abs/2011.13456}
  (\bibinfo{year}{2020}).
\newblock
\showeprint[arXiv]{2011.13456}
\urldef\tempurl%
\url{https://arxiv.org/abs/2011.13456}
\showURL{%
\tempurl}


\bibitem[Sun et~al\mbox{.}(2023)]%
        {sun2023inherently}
\bibfield{author}{\bibinfo{person}{Susu Sun}, \bibinfo{person}{Stefano
  Woerner}, \bibinfo{person}{Andreas Maier}, \bibinfo{person}{Lisa~M. Koch},
  {and} \bibinfo{person}{Christian~F. Baumgartner}.}
  \bibinfo{year}{2023}\natexlab{}.
\newblock \bibinfo{title}{Inherently Interpretable Multi-Label Classification
  Using Class-Specific Counterfactuals}.
\newblock
\newblock
\showeprint[arxiv]{2303.00500}~[cs.CV]


\end{thebibliography}

\end{document}